\documentclass[sigconf]{acmart}
\AtBeginDocument{%
  }

\copyrightyear{2025}
\acmYear{2025}
\setcopyright{acmlicensed}
\acmConference[ICMI '25]{Proceedings of the 27th International Conference on Multimodal Interaction}{October 13--17, 2025}{Canberra, ACT, Australia}
\acmBooktitle{Proceedings of the 27th International Conference on Multimodal Interaction (ICMI '25), October 13--17, 2025, Canberra, ACT, Australia}
\acmDOI{10.1145/3716553.3750773}
\acmISBN{979-8-4007-1499-3/2025/10}

\usepackage{amsmath}
\usepackage{subcaption}
\usepackage{mathtools}

\usepackage{pifont,tikz}

\newcommand\mycopyrighttext{%
\footnotesize \copyright \ Anthony Richardson 2025. This is the author's version of the work. It is posted here for your personal use. Not for redistribution. Please cite the definitive version when referring to this work. The definitive version was published in Proceedings of the 27th International Conference on Multimodal Interaction (ICMI ’25), \url{https://doi.org/10.1145/3716553.3750773}.}

\newcommand\mycopyrightnotice{%
\begin{tikzpicture}[remember picture,overlay]
\node[anchor=north, yshift=-10] at (current page.north) {\fbox{\parbox{\dimexpr\textwidth-\fboxsep-\fboxrule\relax}{\mycopyrighttext}}};
\end{tikzpicture}
}

\begin{document}

\title[Motion Diffusion Autoencoders]{Motion Diffusion Autoencoders: Enabling Attribute Manipulation in Human Motion Demonstrated on Karate Techniques}


\author{Anthony Richardson}
\orcid{0009-0002-0524-8294}
\affiliation{%
  \institution{Cognitive Systems Lab}
  \institution{University of Bremen, Germany}
  \country{}
  }
\email{anthony.richardson@uni-bremen.de}

\author{Felix Putze}
\orcid{0000-0001-5203-8797}
\affiliation{%
  \institution{Cognitive Systems Lab}
  \institution{University of Bremen, Germany}
  \country{}
  }
\email{felix.putze@uni-bremen.de}


\begin{abstract}
  Attribute manipulation deals with the problem of changing individual attributes of a data point or a time series, while leaving all other aspects unaffected. This work focuses on the domain of human motion, more precisely karate movement patterns. To the best of our knowledge, it presents the first success at manipulating attributes of human motion data. One of the key requirements for achieving attribute manipulation on human motion is a suitable pose representation. Therefore, we design a novel continuous, rotation-based pose representation that enables the disentanglement of the human skeleton and the motion trajectory, while still allowing an accurate reconstruction of the original anatomy. The core idea of the manipulation approach is to use a transformer encoder for discovering high-level semantics, and a diffusion probabilistic model for modeling the remaining stochastic variations. We show that the embedding space obtained from the transformer encoder is semantically meaningful and linear. This enables the manipulation of high-level attributes, by discovering their linear direction of change in the semantic embedding space and moving the embedding along said direction. All code and data is made publicly available.
\end{abstract}


\begin{teaserfigure}
  \includegraphics[width=\textwidth]{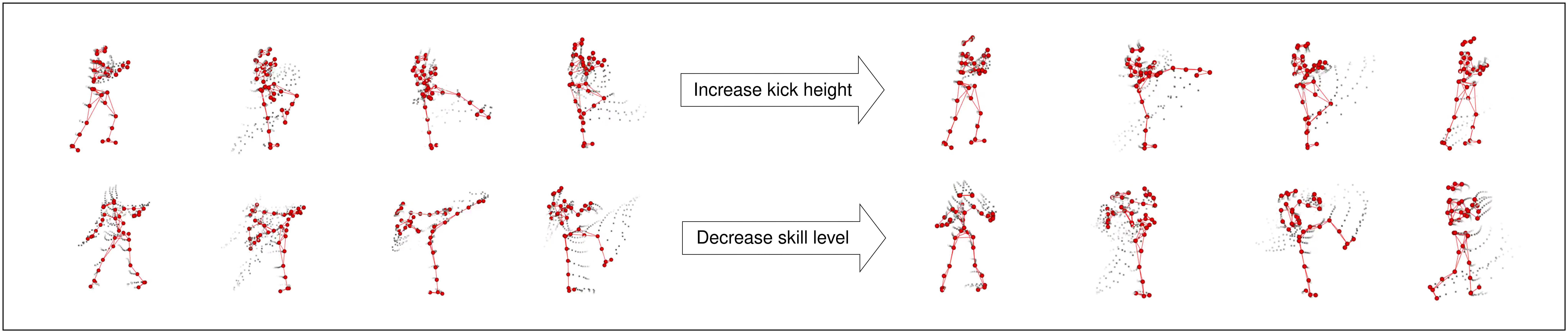}
  \caption{Manipulating the technique and skill level of karate athletes with our proposed Motion Diffusion Autoencoder.}
  \Description{T.}
  \label{fig:teaser}
\end{teaserfigure}


\maketitle


\section{Introduction}
\mycopyrightnotice

Diffusion probabilistic models have driven significant progress in generative modeling \cite{ho2020denoising,song2020denoising,dhariwal2021diffusion}, inspiring a range of new applications. One such application is attribute manipulation \cite{preechakul2022diffusion}, which aims to modify specific attributes of data while preserving others. Applied to human motion, this could open up new possibilities across disciplines: In kinesiology, it could help analyze how diseases or injuries affect movement, aiding in the design of more accessible environments. In sport science, it could serve as a training tool, showing athletes how to refine or adapt their skills. In computer animation, it could eliminate the need for repetitive manual work by generating diverse motions from a single example.

While attribute manipulation has seen success in other domains, such as manipulating facial attributes in images \cite{shen2017learning,akhtar2020utility,wang2018weakly,preechakul2022diffusion}, human motion manipulation is confronted with additional challenges. Firstly, human motion is sequential and its attributes can be \textit{time-variant}. For example, the skill levels of two athletes might be highly distinguishable during the most challenging segments of the executed motion, but show little difference in other parts. Secondly, current state-of-the-art pose representations discard all anatomical details, which is suboptimal for tasks involving attribute manipulation. 
In sports science, for instance, this abstraction results in motion demonstrations based on generic, uniform skeletons that fail to capture the recorded individual's true body proportions. The resulting mismatch can can degrade the user's experience and limit the practical utility of these demonstrations, as viewers may struggle to relate the motions to their own bodies. Therefore, motion demonstrations intended for personalized training should preserve the individual's anatomical structure. We address both challenges mentioned above and present the first success at manipulating attributes of human motion data, in the context of karate techniques.

For such karate motions, we aim to manipulate either the technique or the skill level of the recorded athlete. When changing one of those attributes, the other one should be preserved. Both of these attributes can be regarded as high-level semantics. However, given the stochastic nature of human motion, not all technique executions by athletes, sharing the same semantics, will be the same \cite{bartlett2008movement,aliakbarian2020stochastic}. Those differences can be interpreted as stochastic variations, specific to the athlete or the recording. During manipulation, those stochastic variations should not be lost, so that generated samples are not reduced to few templates. Accordingly, when manipulating a particular attribute, both the remaining semantics and the stochastic variations should be preserved during the process. 

To achieve this preservation, we follow Preechakul et al. \cite{preechakul2022diffusion} and explicitly model the high-level semantics and low-level stochastic variations in two separate embeddings. Manipulation is performed only on the semantic embedding. If the semantic embedding space is linearly separable with regards to the attributes, motion manipulation becomes possible by moving the semantic embedding in the linear direction of change for the specific attribute. Such a manipulation would preserve the remaining semantics of the embedding. As the stochastic embedding is not changed, the aspects captured by it are also maintained. Afterwards, the original stochastic and the new semantic embedding are used to reconstruct a corresponding motion. To support future research on attribute manipulation, all code and data is made publicly available\footnote{The repository can be found at \url{https://github.com/anthony-richardson/MoDiffAE}}.

\section{Related Work}

\textbf{Human Motion Manipulation}: To the best of our knowledge, there exists no system that can manipulate individual high-level attributes in human motion, while preserving the remaining attributes as well as low-level details, unique to the specific motion execution. However, attribute manipulation has seen various success in the image domain. Elarabawy et al. \cite{elarabawy2022direct} introduce \textit{direct inversion}, which models a trade-off between the preservation of the original sample and the realization of the desired manipulation. A diffusion model is expected to preserve the semantic aspects not targeted by the manipulation. However, according to Preechakul et al. \cite{preechakul2022diffusion}, the embedding space of diffusion models lacks such abilities due to their stochastic nature. Instead, they propose to explicitly model the separation of a semantic and a stochastic embedding space. Manipulation is then achieved by moving the semantic image embedding in the linear directions of change for specific attributes. In a work parallel to this, Kim et al. \cite{kim2023diffusion} transfer the idea from Preechakul et al. \cite{preechakul2022diffusion} to the task of face video editing, where the main challenge is to achieve temporal consistency among edited frames. High-level semantics are assumed to be \textit{time-invariant} and are computed frame wise. However, this approach is predetermined to fail on human motion and any other sequence domain, where the semantic aspects require temporal context to be observed and are \textit{time-variant}, meaning their evidence may vary throughout time. Most recently, video diffusion models have demonstrated capabilities of manipulating videos through text guidance. However, they struggle in fields such as human motion, were intricate details and small variations can markedly influence the overall perception \cite{chai2023stablevideo, feng2024ccedit}. We approach this challenge by focusing on human motion capture and explicitly modeling manipulation attributes instead of using text-driven methods. Moreover, we transfer the approach from Preechakul et al. \cite{preechakul2022diffusion} into the sequence domain without restricting the model to \textit{time-invariant} attributes. 

\textbf{Human Pose Representations}: Human pose representation is fundamental to modeling human motion \cite{guo2020action2motion}. A common approach it to represent skeletal poses using 3D Cartesian coordinates \cite{hussein2013human,han2017space}, but this introduces extra barriers in faithful modeling of human kinematics. Constraints that are inherent to the skeleton, such as constant bone length, need to be learned by the model. 
Rotation-based representations, by contrast, disentangle the skeleton structure from its motion trajectory \cite{guo2020action2motion,liu2022investigating}. Many methods adopt continuous rotation representations---free of singularities and differentiable across the full rotation space---to improve training stability and convergence \cite{grassia1998practical,saxena2009learning}. In a quantitative comparison, Liu et al. \cite{liu2022investigating} showed that the continuous, rotation-based Stiefel manifold representation consistently outperforms other representations.
Despite these advantages, existing rotation-based approaches rely on joint rotations of a uniform skeleton and discard anatomical details \cite{guo2020action2motion,tevet2022human,mandery2015kit}. Additional models are then designed to restore the joint coordinates of the uniform skeleton based on the computed rotations \cite{plappert2016kit,pavlakos2018learning}. 
However, since the goal of attribute manipulation is to preserve all aspects except those explicitly targeted by the manipulation, any unintended change in anatomy due to information loss is undesirable. To avoid this, we aim to design a reversible mapping between joint positions and rotations, ensuring that the skeleton of the manipulated motion matches that of the recorded individual. To this end, we propose a novel rotation-based pose representation that leverages the continuous Stiefel manifold to improve training stability and convergence, but also enables the preservation of anatomical details through an iterative reconstruction process.

Beyond the state of the art, the contributions of this paper can be summarized as follows:

\begin{enumerate}
    \item The first success at manipulating attributes of human motion data.
    \item The investigation of the learned embedding space in terms of linearity and semantic interpretability.
    \item The creation of the first continuous, rotation-based pose representation that preserves anatomical details.
\end{enumerate}

\section{The Challenges of Benchmarking Human Motion Manipulation}
\label{chall}
Since there is no prior work that achieves attribute manipulation on human motion data, there exists no benchmark to compare our models performance against. Moreover, when creating such a benchmark, we faced two main difficulties. 

\textbf{Absence of large suitable datasets}: 
For an attribute to be suitable for manipulation, attribute differences should be reflected in the motion capture data. Moreover, at least two suitable attributes are needed in order to evaluate the preservation of untargeted attributes. We further exclude markerless motion capture datasets such as AMASS \cite{mahmood2019amass}, HumanAct12 \cite{guo2020action2motion} and Motion-X \cite{lin2024motion}. On top of achieving inferior recording accuracy, they represent motions using uniform skeletons and thus fail to capture anatomical details. When inspecting the largest remaining datasets, we discovered that most of them, including Human3.6m \cite{ionescu2013human3}, KIT-ML \cite{plappert2016kit} and UESTC \cite{ji2019large}, do not fulfill both criteria mentioned above. To the best of our knowledge, the Kyokushin karate dataset \cite{szczkesna2021optical} is the largest available marker-based motion capture dataset that does so.

\textbf{Lack of reliable metrics for small datasets}: Related work in the field of attribute manipulation performs model evaluation by measuring the  Fr\'echet Inception Distance (FID) between the manipulated samples and test groups that represent the desired attribute changes \cite{preechakul2022diffusion}. A successful manipulation is expected to produce samples with a distribution closest to that of the target attribute. However, to obtain reliable FID measurements, Heusel et al. \cite{heusel2017gans}, recommend a minimum of 10k samples per group. Applying this rule, even the least data demanding scenario, consisting of manipulating two binary attributes, already requires a test set with four groups of 10k samples each. To the best of our knowledge, there neither exists a suitable motion capture dataset large enough to meet this requirement nor an alternative distribution-based metric with a data demand low enough for the existing suitable motion capture datasets. As an alternative to distribution-based approaches, Karras et al. \cite{Karras_2019_CVPR} propose to quantify the latent space disentanglement by measuring its linear separability. We adapt this idea into the domain of human motion manipulation and further inspect the latent space after projection into two dimensions.  

Given these limitations at quantifying the models performance on human motion manipulation, we additionally perform a qualitative evaluation, where we first formulate expected outcomes based on expert knowledge and then inspect whether the test manipulations match those expectations. We use and publish fixed splits of the rigorously preprocessed data, so that the measurements of linear separability and the qualitative evaluation serve as the first benchmark for human motion manipulation. 

\section{Data}

\begin{figure}[t]
    \includegraphics[width=1\columnwidth]{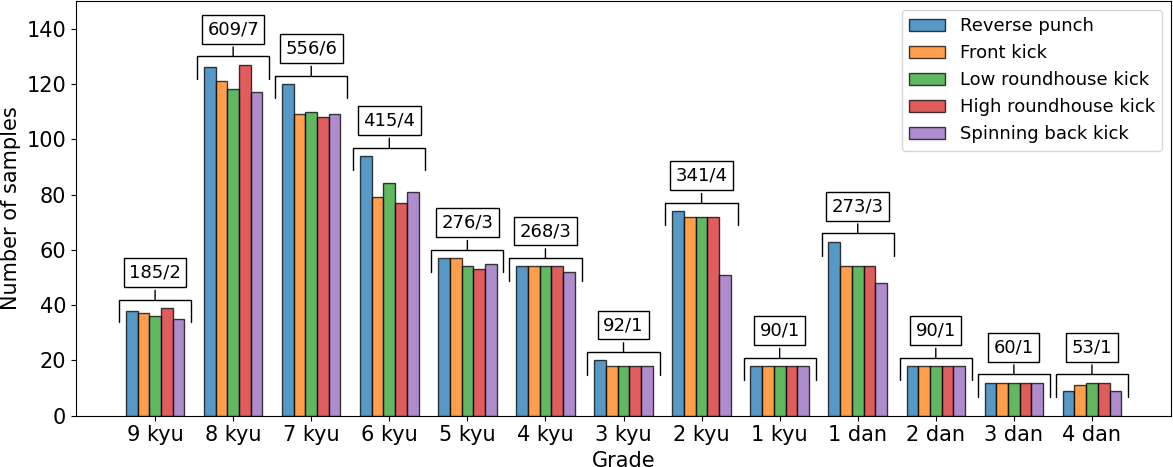}
    \centering
    \caption[Number of samples for each technique per grade]{Number of samples for each technique per grade. The boxes over the bars of each grade give the total number of samples S as well at the number of participants P for that grade. Here, the format is S/P.}
    \label{samplesPerGradePerTechnique}
\end{figure}

We utilize the Kyokushin karate dataset, collected by Szczkesna et al. \cite{szczkesna2021optical}. Thirty-seven healthy individuals, between the ages of 10 and 50, participated in the study. 
Most importantly for this work, the dataset includes two attributes suitable for manipulation. These take the form of the karate grades of the athletes, ranging from 9th kyu (lowest) to 4th dan (highest), and the executed techniques. The study included five different techniques: Reverse punches (RP), spinning back kicks (SBK), front kicks (FK) as well as low (LRK) and high roundhouse kicks (HRK). 
The data was captured using a Vicon motion tracking system \cite{Vicon} at a sampling rate of 250 Hz \cite{szczkesna2021optical}. After our preprocessing steps, explained in Section \ref{preprocessing}, the dataset consists of 3308 samples. The distribution of these samples across the different techniques and grades is shown in Figure \ref{samplesPerGradePerTechnique}. While the samples per technique are evenly distributed within each individual grade, a data imbalance can be observed with regards to the overall grade distribution, as the dataset contains more recordings of beginner than of advanced athletes. 

\section{Preprocessing}
\label{preprocessing}

First, the data was sampled down to 25 Hz. Based on the z-score, we then designed statistical criteria to detect various types of outliers. The detected cases were then inspected and either removed or, if possible, corrected. This included particularly long or short recordings and those that showed little to no activity. Furthermore, untypical head movement was detected that corresponded to participants falling down or adjusting markers. Additionally, poses that were part of the calibration process but unrelated to the techniques were detected and trimmed. All samples were centered to start at the origin of the coordinate system and rotated to initially face the negative y-direction. Additionally, they were normalized so that all techniques are executed with the right limbs. This was achieved by first mirroring the left limb executions on the x-axis and then switching their left and right marker names. However, the marker set from Vicon contains additional markers that are placed between adjacent joints at varying height to distinguish between left and right \cite{Vicon}. This helps the automatic labeling process of the Plug-In Gait software from Vicon, but makes the positioning of the markers asymmetrical, which causes irregularities during the previously mentioned switch of the left and right side. To avoid this, we center these additional markers between their neighboring joints. Detailed formulas for the preprocessing steps can be found in Appendix 1.

\section{Pose Representation}

Accurate attribute manipulation on human motion data calls for a pose representation that preserves anatomical details. However, existing representations either discard such details or are not continuous and thus suffer from stability and convergence issues during training. Thus, we propose the first continuous, rotation-based pose representation that allows the preservation of anatomical details. 

The core idea of our representation is to approximate bone lengths based on the distances of markers that are placed on adjacent joints of the human skeleton. We will refer to such markers as \textit{adjacent markers}. Given the constraint of constant bone lengths in human anatomy, the distance between adjacent markers remains approximately constant throughout the whole motion. In combination with a chain of adjacent markers, defined in Appendix 2, this makes it possible to formulate a reversible mapping from coordinates into a continuous pose representation that preserves anatomical details. 

\subsection{From Coordinates to Features}

\begin{figure}[t]
	\centering
    \includegraphics[width=1\columnwidth]{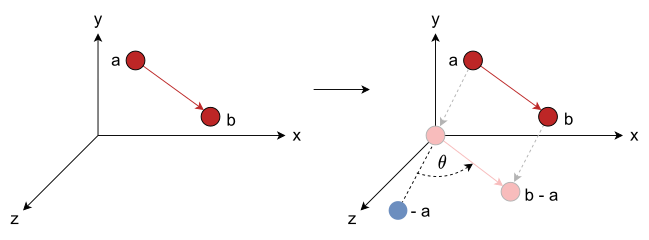}
	\caption{Extracting the angle $\theta$ for two markers $a$ and $b$.}
    \label{fromCoordToFeaturesImage}
\end{figure}

Given the original coordinate representation, an axis angle $\omega$ for two adjacent markers $a$ and $b$ is calculated by
\begin{align}
    &\omega = \underbrace{\frac{a' \times b'}{||a' \times b'||_2}}_{\text{axis } k} \cdot \underbrace{\text{atan2} \left( a' \times b' \cdot \frac{a' \times b'}{||a' \times b'||_2}, a' \cdot b'\right)}_{\text{angle } \theta}, \\
    &\text{where} \quad a' = \frac{-a}{||-a||_2} \quad \text{and} \quad b' = \frac{b-a}{||b-a||_2}
\end{align}
As visualized in Figure \ref{fromCoordToFeaturesImage}, the angle $\theta$, used to compute $\omega$, describes the rotation between the origin of the coordinate system and $b$, where $a$ is the center of rotation. Specifically this angle was chosen, due to its property of allowing a unique reconstruction of $b$. In contrast, the angle between $a$ and $b$ can result in ambiguous cases \cite{dobbsmotivate}. Afterwards, we calculate its rotation matrix $R_k (\theta)$ using the Rodrigues rotation formula and map the result into the Stiefel manifold representation. Said mapping functions by dropping the last column of the rotation matrix \cite{zhou2019continuity}:
\begin{align}
    g_{\text{GS}}\left( \left[ \begin{array}{ccc}
\mid & \mid & \mid \\
r_1 & r_2 & r_3 \\
\mid & \mid & \mid \\
\end{array}\right] \right) = \left[ \begin{array}{cc}
\mid & \mid \\
r_1 & r_2 \\
\mid & \mid \\
\end{array}\right]
\end{align}
To obtain the marker distance between $a$ and $b$, we calculate the average distance $d$ over all time steps of the sample. The resulting distance is treated as a constant and not passed through the model. These steps are repeated for each link in the chain of adjacent markers. 

\subsection{From Features to Coordinates}

The mapping $f_{\text{GS}}$ from the Stiefel manifold representation back to the rotation matrix $R_k (\theta)$ is defined by the following Gram-Schmidt-like process \cite{zhou2019continuity}:
\begin{align}
    &f_{\text{GS}}\left( \left[ \begin{array}{cc}
\mid & \mid \\
r_1 & r_2 \\
\mid & \mid \\
\end{array}\right] \right) = \left[ \begin{array}{ccc}
\mid & \mid & \mid \\
r_1^* & r_2^* & r_3^* \\
\mid & \mid & \mid \\
\end{array}\right] \quad \text{, where} \\
& r_i^* = \left[  \Biggl\{\begin{array}{lr}
    N(r_1) & \text{if } i = 1 \\
    N(r_2 - (r_1^* \cdot r_2)r_1^*) & \text{if } i = 2 \\
    r_1^* \times r_2^* & \text{if } i = 3 
\end{array}  \right]^T
\end{align}

$N(\cdot)$ denotes a normalization function, which divides each vector component by the length of the vector so that all components add up to one. Once the matrix $R_k (\theta)$ is obtained, its axis and angle of rotation can be extracted. The angle $\theta$ is given by 
\begin{align}
    \theta = \arccos \left( \frac{\text{trace}(R_k (\theta)) - 1}{2} \right)
\end{align}
where $\text{trace($\cdot$)}$ refers to the sum of the diagonal elements of the matrix. Moreover, knowing that the axis of rotation is an eigenvector of $R_k(\theta)$, corresponding to an eigenvalue of 1, it can be found by solving $(R_k (\theta) - I)k = 0$ towards $k$. Given the position of an adjacent marker $a$, the stored distance between $a$ and $b$, as well as the axis $k$ and angle $\theta$, it then becomes possible to reconstruct the coordinates of $b$ by
\begin{align}
    & b = \text{rod}\left( \frac{k}{||k||_2}, \theta, \frac{-a}{||-a||_2}  \right) \cdot d_{ab} + a \quad \text{, where} \\
    & \text{rod}(u,  \theta, v) = v \cos(\theta) + (u \times v) \sin(\theta) \\ & \quad \quad \quad \quad \; \: \, + u (u \cdot v) (1 - \cos(\theta))
\end{align}
This concludes the reconstruction of $b$. When done iteratively according to the defined chain of adjacent markers, it becomes possible to reconstruct the whole skeleton. The position of the first marker in the chain can be chosen arbitrary or, when translation is included, according the coordinates of one marker that is passed through the model in addition to the angles. Moreover, the explained reconstruction process was designed to be differentiable. During the training of neural networks, it thus becomes possible to incorporate loss functions that utilize joint positions. We make use of this property in Section \ref{objective}.

\subsection{Assumption of Constant Joint Distances}
\label{assumption}

In the proposed reversible mapping, the average distances between adjacent joints are calculated and treated as constants. In order to assess whether this assumption is justified, the extent to which the actual distance between the respective joints changes during the entire movement is analyzed. We determine that the average standard deviation of the joint distances is 0.35 cm and therefore consider the assumption of constant distances to be justified. Nonetheless, larger errors can potentially accumulate during the iterative reconstruction of the skeleton. To analyze how large this accumulated error is, we transform the joint positions of the whole dataset into the designed pose representation and then reconstruct them using the presented approach. This results in an average reconstruction error of 1.05 cm. We argue that such a small reconstruction error is negligible in the context of full-body motion, as it falls within natural human movement variability and is imperceptible during dynamic actions. In particular, this reconstruction error pales in comparison to the error caused by other existing rotation-based pose representations that are unable to preserve anatomical details.

\section{Architecture}
The embedding space of diffusion models is in itself highly stochastic and not semantically rich, given the fact that representations $x_t$ are obtained by adding Gaussian noise. This makes their embedding spaces unsuitable for attribute manipulation, despite their ability to accurately reconstruct noisy samples \cite{preechakul2022diffusion}. The proposed architecture aims at creating a semantically rich embedding space that enables attribute manipulation on sequential data, while maintaining the high reconstruction capabilities of diffusion models. In other words, it seeks to extract a semantically meaningful and decodable representation. Preechakul et al. \cite{preechakul2022diffusion} argue that this requires capturing both the high-level semantic and low-level stochastic variations. To obtain a semantically meaningful and decodable representation, we design an autoencoder, consisting of three components: A stochastic encoder to produce the stochastic embedding $x_T$, a semantic encoder to produce the semantic embedding $z$ and a decoder to reconstruct the original motion $x_0$ based on those embeddings. We refer to the reconstructed motion as $\hat{x}_0$. An overview of the architecture is shown in Figure \ref{fig:baseConfig}. The detailed architecture and the chosen hyperparameters can be found in Appendix 4. As this system is an adaptation of the Diffusion Autoencoder by Preechakul et al. \cite{preechakul2022diffusion} into the motion domain, we call our model Motion Diffusion Autoencoder, or MoDiffAE in short. 

\begin{figure}[t]
	\centering
	\includegraphics[width=1\columnwidth]{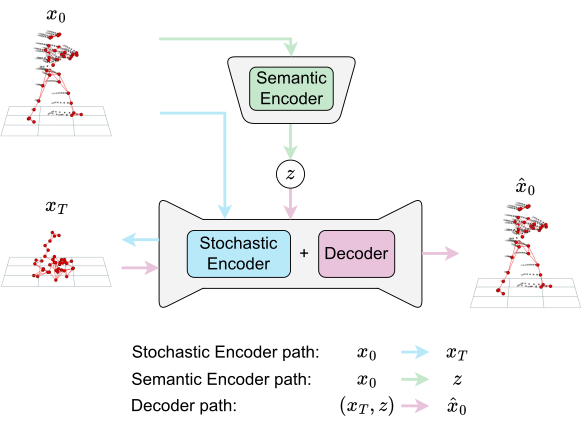}
	\caption{Overview of the MoDiffAE architecture.}
	\label{fig:baseConfig}
\end{figure}

\subsection{Decoder}
We design a Denoising Diffusion Implicit Model (DDIM) which models $p_\theta(x_{t-1} | x_t, z)$ and is conditioned on an additional latent variable $z$, representing the semantic embedding. Since the embedding space of diffusion models, given by the $x_t$ at different diffusion steps $t$, struggles to capture semantic aspects, it will be regarded as stochastic embedding. The decoder then starts at $x_T$ and uses the model $p_\theta(x_{t-1} | x_t, z)$ to iteratively predict $\hat{x}_0$, following the Markov chain of the reverse diffusion process. This is defined as
\begin{align}
    q_\theta(x_{0:T} | z) &= q(x_T) \prod_{t=1}^{T} p_\theta (x_{t-1} | x_t, z) \quad \text{, where} \\
    q(x_T) &= \mathcal{N}(x_{T}; 0, 1)
\end{align}

The model $p_\theta(x_{t-1} | x_t, z)$ first predicts $\hat{x}_0$ and then diffuses it back to $x_{t-1}$. This is defined as
\begin{align}
    p_\theta(x_{t-1} | x_t, z) &= \sqrt{\overline{\alpha}_{t-1}} \hat{x}_0 + \sqrt{1 - \overline{\alpha}_{t-1}}\epsilon
    \label{prediciontAndDiffusion} \quad \text{, where} \\
    \hat{x}_0 &= m_\theta(x_t, t, z) \quad \text{and} \quad \epsilon \sim \mathcal{N}(x_{t}; 0, 1)
\end{align}
The different $\overline{\alpha}$ are given by the employed noise schedule and $m_\theta$ denotes a transformer encoder that predicts $\hat{x}_0$. 

\subsection{Semantic Encoder}

\begin{figure*}[ht]
    \subfloat{
        \centering
        \includegraphics[width=1\columnwidth]{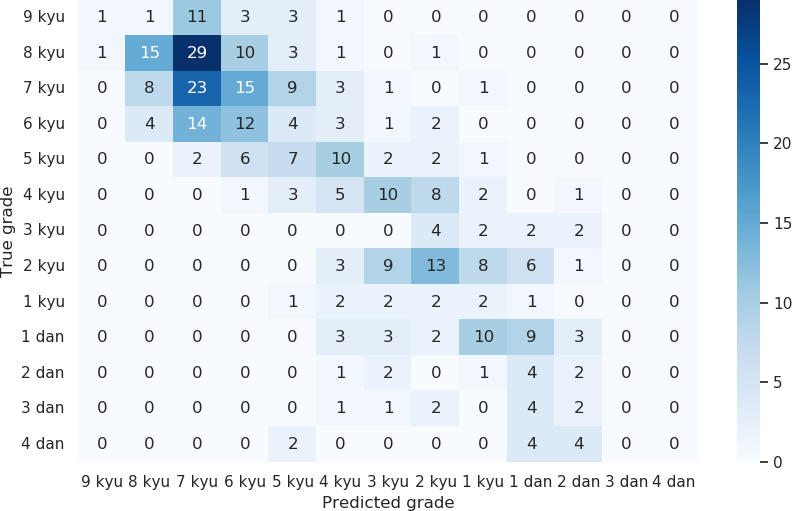}
        \label{regressionConfMatrixGradeB0401Validation}
    }
    \subfloat{
        \centering
        \includegraphics[width=1\columnwidth]{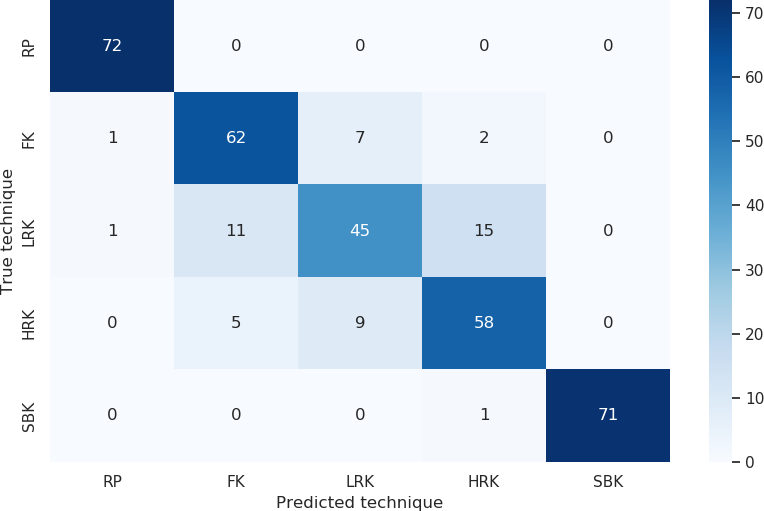}
        \label{regressionConfMatrixTechniqueB0401Validation}
    }
    \caption{Grade (left) and technique (right) confusion matrices for validation data.}
    \label{confusionMat}
\end{figure*}

\begin{figure*}[ht]
    \hspace{0.47cm}
    \subfloat{
        \centering
        \includegraphics[width=\columnwidth]{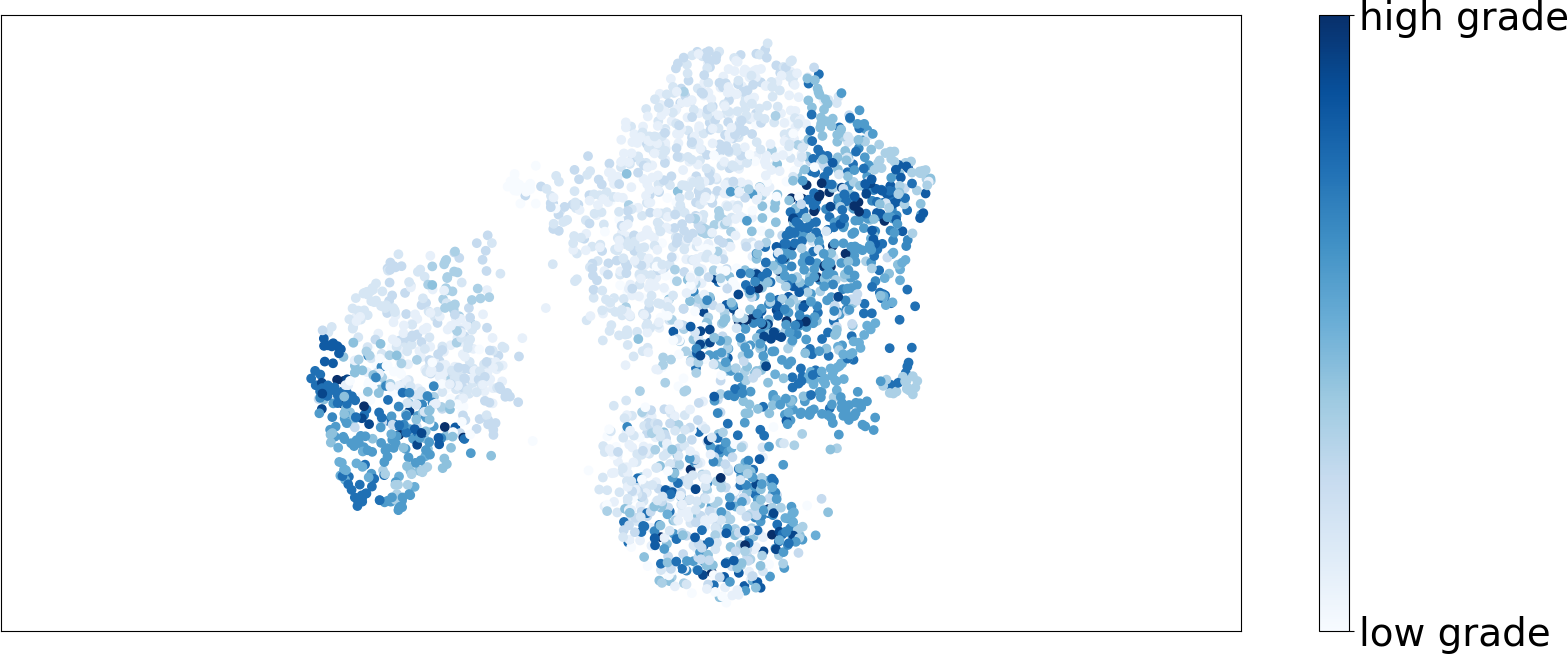}
        \label{gradeEmbeddingB0401Train}
    }
    \hspace{0.2cm}
    \subfloat{
        \centering
        \includegraphics[width=0.4\textwidth]{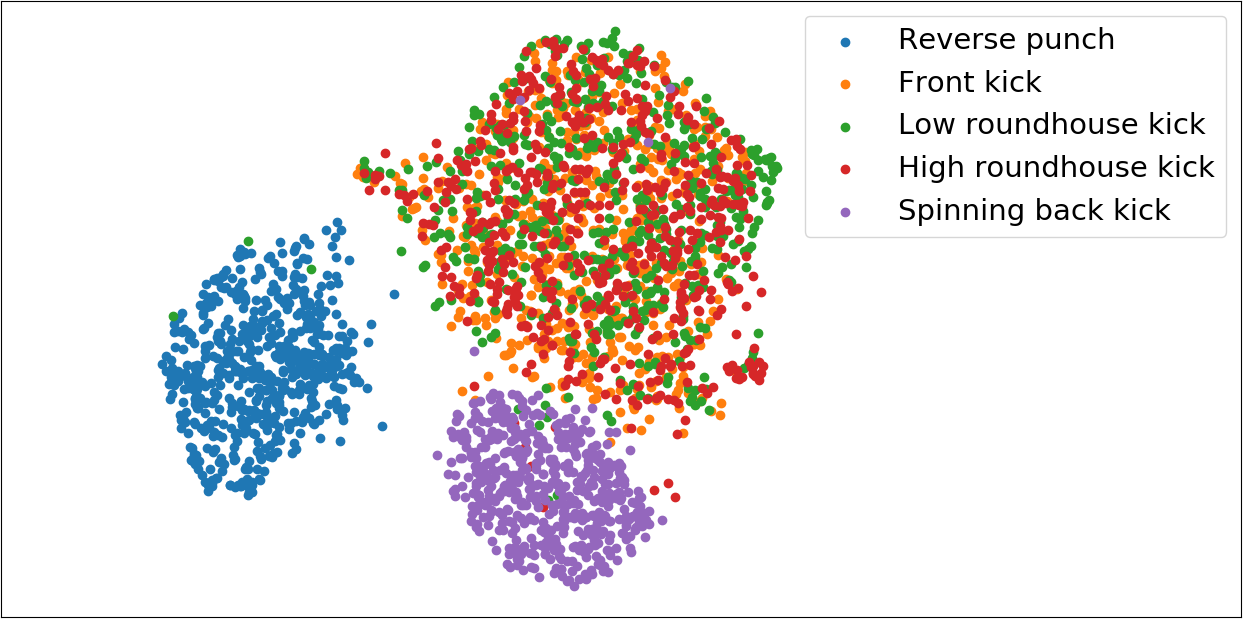}
        \label{techniqueEmbeddingB0401Train}
    }
    \caption{UMAP projection of semantic embeddings of the train data with visualized grades (left) and techniques (right).}
    \label{umapPlots}
\end{figure*}

An additional transformer encoder $f_\psi(x_0) = z$ is designed to take the original motion $x_0$ as input and produce the corresponding semantic embedding $z$, needed by the decoder. The architecture is built in a way that motivates $f_\psi(x_0)$ to capture information, that assists the decoder in the reconstruction of the original motion and that the stochastic embedding $x_T$ fails to capture. 
Given that diffusion models excel at capturing stochastic details but lack semantic richness of their embedding space, the model $f_\psi(x_0)$ will be incentivized to capture those missing semantic aspects. From an optimization perspective, there is little sense for the semantic encoder to capture stochastic aspects if the stochastic encoder is much better at doing so. While there is no guarantee that this will result in purely semantic and stoachstic embeddings, success was already demonstrated in the image domain \cite{preechakul2022diffusion}.  

\subsection{Stochastic Encoder}

The model $m_\theta$, used by the decoder, can also be used in a deterministic diffusion process to encode an input motion $x_0$ into the corresponding stochastic embedding $x_T$. In the context of $m_\theta$ and wrapped into a recursive function, this process is defined as
\begin{align}
    h_\theta(x_t) &= \Biggl\{\begin{array}{lr}
    x_t & \text{if } t = T \\
    b_\theta(h(x_t)) & \text{else} 
    \end{array} \quad \text{, where} \\
    \label{detReverseModel}
    b_\theta(x_t) &\approx \sqrt{\overline{\alpha}_t} \hat{x}_0 + \sqrt{1 - \overline{\alpha}_t} \hat{\epsilon} \\
    \hat{x}_0 &= m_\theta(x_{t-1}, t-1, z) \\
    \hat{\epsilon} &= \frac{x_{t-1} - \sqrt{\overline{\alpha}_{t-1}} m_\theta(x_{t-1}, t-1, z)}{\sqrt{1 - \overline{\alpha}_{t-1}}}
\end{align}

Here, $b_\theta$ denotes the process that deterministically diffuses $x_{t-1}$ to $x_{t}$. This deterministic process is not used during training, but only during the guided manipulation, explained in Section \ref{guidedManipulation}. 

\subsection{Training Objective}
\label{objective}

The first component of the loss function is the \textit{simple loss}, introduced by Ho et al. \cite{ho2020denoising}. When directly predicting $\hat{x}_0$, this loss is defined as: 
\begin{align}
    L_{\text{simple}} \coloneqq& \mathbb{E}\left[|| x_0 - \hat{x}_0 ||_2^2\right] \quad \text{, where} \\
    \hat{x}_0 =& m_\theta(\sqrt{\overline{\alpha_t}} x_0 + \sqrt{1 - \overline{\alpha}_t}\epsilon, t, f_\psi(x_0))
\end{align}

Following Tevet et al. \cite{tevet2022human}, we directly predict $\hat{x}_0$ in order to incorporate domain knowledge in the form of geometric losses, which are defined as
\begin{align}
    L_{\text{pos}} \coloneqq &\mathbb{E} \left[ \frac{1}{L - 1} \sum_{i=0}^{L - 1}  || JP(x_0^i) - JP(\hat{x}_0^i) ||_2^2\right] \label{posLoss}\\
    L_{\text{foot}} \coloneqq &\mathbb{E}\left[ \frac{1}{L - 2} \sum_{i=0}^{L - 2}  || JP(\hat{x}_0^{i+1}(F)) - JP(\hat{x}_0^{i}(F)) \cdot c_i ||_2^2\right] \label{footLoss} \\
    L_{\text{vel}} \coloneqq &\mathbb{E}\left[ \frac{1}{L - 2} \sum_{i=0}^{L - 2}  || (x_0^{i+1} - x_0^{i}) - (\hat{x}_0^{i+1} - \hat{x}_0^{i}) ||_2^2\right] \label{velLoss} 
\end{align}

Here, $JP$ is the differentiable reconstruction process, explained in Section \ref{assumption}, that restores the joint positions based on $x_0$ and the stored distances of neighboring markers. Equation \ref{posLoss} represents the positional loss, penalizing differences between the original and reconstructed joint positions. Equation \ref{footLoss} defines the foot contact loss, where $(F)$ indicates that only the foot markers are used and $c_i \in \{0, 1\}^{|F|}$ is the binary foot contact mask, indicating, for each frame $i$, whether the foot markers in $F$ touch the ground. Following Shi et al. \cite{shi2020motionet}, $c_i$ is set according to binary ground truth data. In essence, $L_{\text{foot}}$ penalizes foot-sliding by nullifying velocities when touching the ground. Lastly, Equation \ref{velLoss} penalizes velocity differences. The overall training loss is defined as 
\begin{align}
    L = L_{\text{simple}} + \phi_{\text{pos}} L_{\text{pos}} + \phi_{\text{foot}} L_{\text{foot}} + \phi_{\text{vel}} L_{\text{vel}} 
\end{align}
where the different $\phi$ are weights, determining the influence of the corresponding components. Training is done by optimizing $L$ with respect to $\theta$ and $\psi$.

\section{Guided Manipulation}
\label{guidedManipulation}

Similar to Preechakul et al. \cite{preechakul2022diffusion}, we train an additional linear layer to predict attributes based on semantic embeddings. More precisely, it is used to predict probabilities for the five karate techniques as well as an estimation of the grade. The technique labels are one-hot encoded, while the grades are modeled continuously on a linear scale between zero and one. During training, the respective MoDiffAE model is frozen. A well performing linear classifier implies that its weights for the individual attributes represent linear directions of change. Accordingly, attribute manipulation becomes possible by moving semantic embeddings in said directions, where $\lambda$ is a factor determining the manipulation strength. While Preechakul et al. \cite{preechakul2022diffusion} choose a fixed $\lambda$ for all manipulations, we extend this approach by a guidance mechanism. When continuously increasing $\lambda$ we found that the attribute predictions based on the resulting embeddings reach a point of convergence, which we term $\lambda_{\text{max}}$. Furthermore, the ideal $\lambda$ w.r.t. to the attributes appears to be at a point between $\lambda=0$ and $\lambda_{\text{max}}$. To determine an appropriate $\lambda$, we therefore interpolate in small steps between those borders, predict the respective attributes using the linear classifier and score all resulting embeddings by equally weighing the distance of the predicted technique and grade to the targets. The embedding with the lowest average distance is chosen as the manipulated semantic embedding. In summary, we first obtain the semantic and stochastic embeddings for a motion, then manipulate the semantic embedding using the explained guidance mechanism and finally reconstruct a motion based on the original stochastic and the manipulated semantic embedding. 

\section{Evaluation}

\definecolor{green(html/cssgreen)}{rgb}{0.0, 0.5, 0.0}
\definecolor{indiagreen}{rgb}{0.07, 0.53, 0.03}

\begin{table*}[t]
  \centering
  \caption{FIDs between test samples and reference groups using semantic embeddings when changing a high roundhouse kick into a front kick. The karate grade of the participant is the 8th kyu. The FIDs between the technique groups are shown on the left. The FIDs between the grade groups are shown on the right. Not listed grades also have larger FIDs.}
  \subfloat{
    \centering
    \resizebox{0.48\textwidth}{!}{
    \begin{tabular}{l|ccccc}
    \hline
        Technique & RP & FK & LRK & HRK & SBK \\ \hline 
        FID before & 87.70 & 33.16 & 29.54 & $\boldsymbol{\textcolor{indiagreen}{26.69}}$  & 55.50 \\ 
        FID after & 115.63 & $\boldsymbol{\textcolor{indiagreen}{85.84}}$ & 87.62 & 86.71 & 127.94 \\ \hline
    \end{tabular}
    }
  }
  \subfloat{
    \centering
    \resizebox{0.49\textwidth}{!}{
    \begin{tabular}{l|ccccc}
    \hline
        Grade & 9 kyu & 8 kyu & 7 kyu & 6 kyu & 5 kyu \\ \hline 
        FID before & 81.40 & $\boldsymbol{\textcolor{indiagreen}{26.69}}$ & 34.18 & 62.22 & 66.55 \\
        FID after & 130.01 & $\boldsymbol{\textcolor{indiagreen}{85.84}}$ & 100.08 & 91.83 & 161.93 \\ \hline
    \end{tabular}
    }
  }
  \label{fids}
\end{table*}

\begin{figure*}[t]
    \centering
    \subfloat[From high roundhouse kick to front kick. (1st dan)]{
        \includegraphics[width=\textwidth]{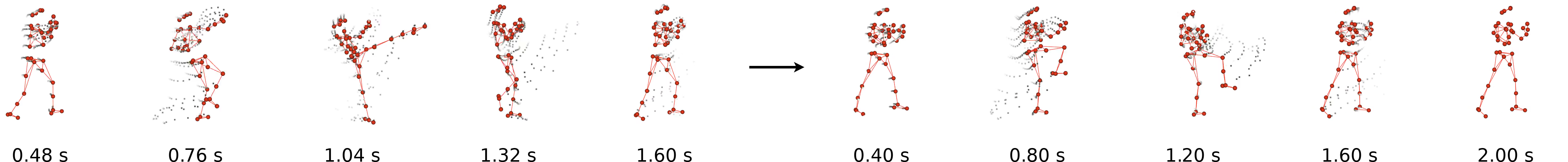}
    	\label{fig:technique_from_high_roundhouse_kick_to_front_kick_3}
    }
    
    \subfloat[From low roundhouse kick to high roundhouse kick. (8th kyu)]{
        \includegraphics[width=\textwidth]{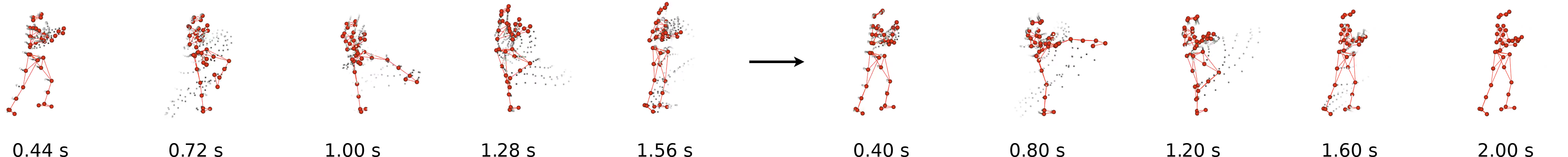}
    	\label{fig:technique_from_low_roundhouse_kick_to_high_roundhouse_kick_7}
    }
    
    \subfloat[From 8th kyu to 1st dan. (reverse punch)]{
        \includegraphics[width=\textwidth]{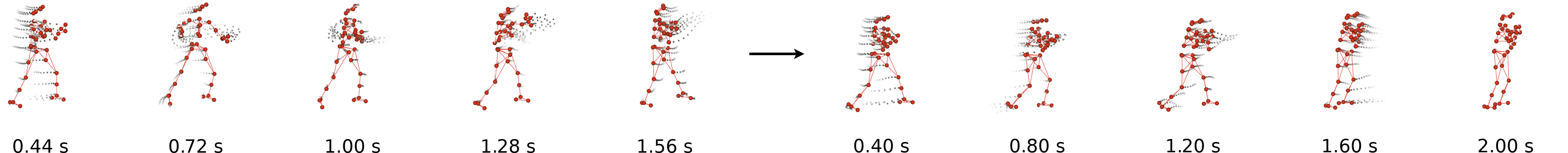}
    	\label{fig:grade_increase_reverse_punch_15}
    }
    
    \subfloat[From 1st dan to 8th kyu. (spinning back kick)]{
        \includegraphics[width=\textwidth]{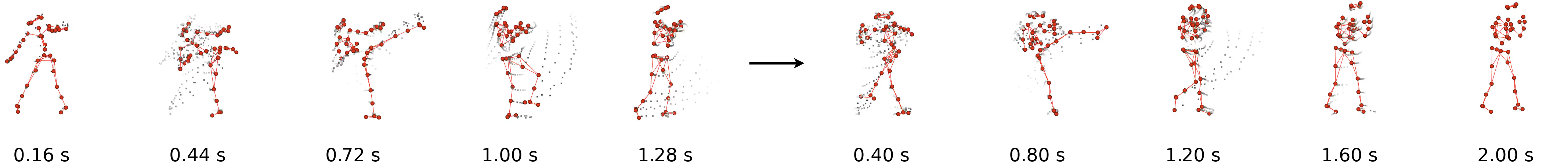}
    	\label{fig:grade_decrease_spinning_back_kick_36}
    }
    \caption{Exemplary technique and grade manipulations. Timestamps are provided below each frame.}
    \label{transAttention}
\end{figure*}

We evaluate our model both quantitatively and qualitatively. The quantitative evaluation analyzes the model’s embedding space, complemented by an FID-based assessment. For the qualitative evaluation, we define expected movement characteristics based on expert knowledge and examine whether these are reflected in the manipulated test samples. In Appendix 6, we further analyze the preservation of stochastic variations across multiple test cases.

\subsection{Quantitative Evaluation}
\label{linearity}

\textbf{Analysis of the Embedding Space}:
Following Karras et al. \cite{Karras_2019_CVPR} and Preechakul et al. \cite{preechakul2022diffusion}, we quantitatively evaluate our model by measuring the linear separability of its embedding space w.r.t. different attributes. This analysis is of special interest for the proposed manipulation approach, as it explicitly relies on a classifier to discover linear directions of change in the embedding space. Figure \ref{confusionMat} shows how said linear classifier performs at predicting the techniques and grades for the validation data. The unweighted average recall of the technique predictions is 0.789. Regarding the skill level, the unweighted mean absolute error is 0.146, which can be interpreted as 1.752 out of 13 grades. It can also be observed that the grade predictions become less accurate at higher grades, which we attribute to the lower number of participants with high skill levels. Nonetheless, the model is in most cases capable of differentiating between beginner and advanced athletes. To further investigate the embedding space, we use Uniform Manifold Approximation and Projection (UMAP) to obtain a 2D representation for the semantic embeddings of the training data. Low-dimensional projections offer limited insight into the linear separability of the embedding, as such separability may only exist in the original high-dimensional space. However, they can provide complementary insights into the embedding structure. When inspecting the projections in Figure \ref{umapPlots}, it can be observed that the reverse punch and spinning back kick form separate clusters, while the remaining kick variations appear very intertwined. This makes sense given how unique the reverse punch and spinning back and how similar the remaining techniques are. Regarding the grades, the 2D embedding space shows a separation of low and high grades inside of each technique cluster. Overall, this analysis demonstrates that, despite being obtained through unsupervised training, the semantic embedding space captures attribute-specific information and is approximately linear separable with regards to those attributes.

\textbf{Exemplary FID-based analysis}: 
Table \ref{fids} shows the FID scores for the techniques and grades when changing from high roundhouse kicks to front kicks. We choose this example as it has the most available samples in the compared groups. Aligning with the intention of the manipulation, it can be seen that the test samples are closest to the expected techniques before and after manipulation, while the closest grade reference group remains the same. However, these measurements should be interpreted with caution. Due to the lack of data, each underlying high dimensional distribution is approximated using less than 50 samples. This results in unreliable measurements and motivates us to complement this with alternative evaluation approaches.

\subsection{Qualitative Evaluation}

Research in martial arts revealed that speed, power, flexibility as well as upper and lower limb synchronization are differentiating factors of different skill levels \cite{chaabene2012physical,chaabene2019needs,8708247,probst2007acomparison}. Concerning the differences between the five techniques, we follow the descriptions from Szczkesna et al. \cite{szczkesna2021optical}. Accordingly, front kicks are executed frontal, whereas roundhouse kicks follow a circular motion and strike from the side. The two roundhouse kick variations differ in their height. More precisely, low roundhouse kicks are performed at knee to hip height, while high roundhouse kicks are targeted at the shoulder or head. The spinning back kick is the only technique including a spin, while the reverse punch is the only technique that is not a kick. We investigated for all test samples whether these skill and technique related characteristics are changed or preserved during manipulation. The examples shown in Figure \ref{transAttention} represent frequent observations in the test manipulations. Further test manipulations are visualized in Appendix 7. 

\textbf{Technique Manipulations}: We observe that the models create plausible and realistic motions when changing between the front kicks as well as the low and high round house kicks. Moreover, clear signs of grade preservation are noticeable. Figure \ref{fig:technique_from_high_roundhouse_kick_to_front_kick_3} shows a technique manipulation from a high roundhouse to a front kick. After manipulation, the kick is performed completely frontal, i.e., without any rotation. Regarding the grade preservation, we first inspect the time difference of the right foot starting to move and it returning to the ground. It can be seen that the duration, and therefore the speed of the kick stays similar after manipulation. Secondly, the arms are swung synchronously in the opposite direction of the kick in both cases, presumably to generate power. Furthermore, Figure \ref{fig:technique_from_low_roundhouse_kick_to_high_roundhouse_kick_7} shows a manipulation from a low to a high roundhouse kick. Here, the rotation is maintained, while the kick height is increased. Again, the speed of the kick as well as the amount of upper and lower limb synchronization stay similar throughout the manipulation. In contrast, we find the models to struggle with technique manipulations that involve reverse punches and spinning back kicks. In many of those test cases, the type of technique is not changed. 

\textbf{Grade Manipulations}: 
We observe that the models perform grade manipulations that align with the skill factors previously mentioned, while preserving the type of technique. Figure \ref{fig:grade_increase_reverse_punch_15} shows a grade increase for a reverse punch. It can be seen that the speed and therefore also the power of the punch is increased. Originally, the duration, between the start of the first arm swing at 0.44 s and the return to a neutral position at 1.56 s, is 1.12 s. After manipulation, this time difference is reduced to roughly 0.8 s. Figure \ref{fig:grade_decrease_spinning_back_kick_36} shows a grade decrease for a spinning back kick. Originally, the kick exceeds head height and is executed with two straight legs, requiring a high level of flexibility. After decreasing the grade, the character does not reach the same kicking height, even when bending both legs, indicating a lower level of flexibility. Additionally, the arm swing in the opposite direction of the kick is less noticeable after manipulation, which is tied to power as well as upper and lower limb synchronization.

\section{Conclusion}

To the best of our knowledge, Motion Diffusion Autoencoders present the first success at attribute manipulation on human motion data. They demonstrate successful technique manipulations on three out of five techniques and are capable of changing the skill level from low to high and vice versa. Moreover, we propose the first benchmark on human motion manipulation. Distribution-based evaluation approaches for attribute manipulation require a test set, containing reference groups for different attributes and their values. In domains like human motion, where high-quality data is limited, these groups are often small, which undermines the reliability of approximating the underlying high-dimensional distributions. Therefore, we adopt evaluation techniques that provide reliable insights despite the comparatively small size of motion capture datasets. Nevertheless, existing distribution-based evaluation metrics remain unreliable when data is limited, which highlights two promising directions for future research: advancing evaluation methods for attribute manipulation on small datasets or constructing larger, more balanced datasets to improve the reliability of existing distribution-based evaluation approaches.

The manipulation of human motion data was in part enabled by the design of a novel continuous, rotation-based pose representation. We showed that it enables the disentanglement of the human skeleton and its motion trajectory, while still allowing an accurate reconstruction of the original anatomy. However, this representation is not directly applicable to use-cases with different markers sets. Instead, one needs to design a new chain of adjacent markers and re-evaluate whether the assumption of constant distances is justified for that particular chain. In the future, it could be attempted to design a pose representation that has the same qualities but is independent of the specific marker set.

\section{Safe and Responsible Innovation Statement}

The proposed system offers significant benefits across healthcare, sports science, and animation. However, its deployment raises important ethical considerations. Ensuring data privacy through informed consent and anonymization is critical, especially given the identifiable nature of motion data. The imbalance in training data highlights the need for inclusive, diverse datasets to avoid biased outcomes, particularly for underrepresented populations. While not in the scope of this work, attempts could be made to extend the presented methods to markerless motion capture in videos. Once the current limitations of video diffusion models for motion generation have been overcome, the corresponding systems could be misused to fabricate realistic but false video evidence.

\section*{Acknowledgments}
This work was funded by the Deutsche Forschungsgemeinschaft (DFG, German Research Foundation) - 459360854 as part of the Research Unit "Lifespan AI: From Longitudinal Data to Lifespan Inference in Health" (DFG FOR 5347) (http://lifespanai.de).


\bibliographystyle{ACM-Reference-Format}
\bibliography{modiffae}

\end{document}